\documentclass[sigconf]{acmart}
\AtBeginDocument{%
  }

\copyrightyear{2026}
\acmYear{2026}
\setcopyright{cc}
\setcctype{by}
\acmDOI{XXXXXXX.XXXXXXX}
\acmConference[WCCCE 2026]{Western Canada Conference on Computing Education 2026}{April 30--May 1,
  2026}{Vancouver, Canada}
  
\acmISBN{978-1-4503-XXXX-X/2026/06}

\usepackage{todonotes}
\usepackage{booktabs}
\usepackage{subcaption}

\usepackage{tikz}
\usetikzlibrary{shapes.arrows, positioning, calc}

\begin{document}

\title{AI-Generated Slides: Are They Good? Can Students Tell?}

\author{Juho Leinonen}
\orcid{0000-0001-6829-9449}
\affiliation{
  \institution{Aalto University}
  \city{Espoo}
  \country{Finland}
}
\email{juho.2.leinonen@aalto.fi}
\authornote{Equal contribution. Author order is alphabetical, which coincidentally favours the first author.}

\author{Lisa Zhang}
\orcid{0000-0002-7302-6530}
\affiliation{%
  \institution{University of Toronto Mississauga}
  \city{Mississauga}
  \state{Ontario}
  \country{Canada}
}
\email{lczhang@cs.toronto.edu}
\authornotemark[1]

\author{Arto Hellas}
\orcid{0000-0001-6502-209X}
\affiliation{
  \institution{Aalto University}
  \city{Espoo}
  \country{Finland}
}
\email{arto.hellas@aalto.fi}

\renewcommand{\shortauthors}{Leinonen, Zhang, and Hellas}

\begin{abstract}
As generative AI (GenAI) tools become easily accessible,
there is promise in using such tools to support instructors.
To that end, this paper examines using GenAI to help generate 
slides from instructor authored course notes, emphasizing
instructor and student perceptions.
We examine an end-to-end education tool (NotebookLM), two general-purpose LLMs (Claude, M365 Copilot), and two coding assistants (Cursor, Claude Code).
We first analyze whether GenAI generated slides are ``good''
via narrative assessment by educators.
We choose the best slides to use (with some modification)
in a real course setting, and
compare the student perception of human vs. AI generated slides.
We find that coding assistant tools produce slides that were
most accurate, complete, and pedagogically sound.
Additionally, students rate GenAI slides 
to be of similar quality as instructor-created slides,
and cannot reliably identify which slides are AI-generated. Additionally, we find a 
negative correlation between a high quality rating and a high ``AI-generated'' rating, suggesting students associate poor quality with the source of the slides being AI.
These findings highlight promising opportunities for integrating GenAI into instructional design workflows and call for further research on how educators can best harness such tools responsibly and effectively.

\end{abstract}





\maketitle


\section{Introduction}

As large language models (LLMs) and AI-powered educational tools rapidly advance,
LLM and Generative Artificial Intelligence (GenAI) tools have been shown to create educational content~\cite{wang2024large}.
They are perceived to increase productivity on both programming tasks~\cite{peng2023impact} and writing tasks~\cite{noy2023experimental}.
However, even within software engineering, there is variation in users' perception~\cite{dakhel2023github}, with GenAI tools sometimes generating incorrect or hallucinated outputs. 
Computing instructors thus face both opportunities and uncertainties in integrating GenAI into their pedagogical practice. Student perceptions of AI-generated materials introduce further complications~\cite{li2024does}.


This paper presents a case study using GenAI tools in
assisting instructors in generating slides from
instructor-authored online textbook chapters,
in the context of an upper-year undergraduate web programming course.
We are interested in two questions:

\textbf{(1) Are they good?} Here, ``they'' refers holistically to
the GenAI tools themselves, the process of using them,
and the generated slide outputs. 
We explore and compare various types 
of GenAI tools:
end-to-end education tools that students may already use (e.g., NotebookLM),
general purpose LLMs that are accessible (e.g., Claude, M365 Copilot), and
coding assistants that have been shown to be helpful for programming tasks (e.g., Cursor, Claude Code).
We report on the instructor experience and their narrative assessment of factual accuracy, completeness, and pedagogical soundness.
These factors are similar to  work conducted at a similar time frame~\cite{xie2026slidebot}.

\textbf{(2) Can students tell?}
We select the best slides generated across tools and, with some modifications, evaluate their quality in a real-world classroom;  we then report student perception of quality and their ability to identify AI-generated content.
Specifically, after matching the visual styling of GenAI-generated slides to the instructor's preferred format and selectively deleting content, we use these slides in actual course lectures alongside instructor-created slides.
Students are informed that lectures may contain GenAI-generated content but are not told which segments or how many.
After each lecture segment, students rate the slide quality 
and guess whether the slides were AI or human generated.



We choose tools that are accessible (free, <USD\$30 per month, or available via institutional licenses). 
We prioritize tools that have not been widely discussed in the computing education literature~\cite{prather2025beyond,prather2023robots},
to introduce new tools to the community.
Additionally, slides are a common form of content used to support lectures that is time-consuming to create.
Slides are also challenging for GenAI tools as its creation involves 
coordination across text, code, and diagrams.

As the authors are computing educators and researchers with minimal (<2 months) prior experience using these specific GenAI tools to build educational resources, we offer an authentic perspective on the instructor on-boarding experience. 
Additionally, one researcher serves as the instructor for the course examined,
providing domain expertise for evaluating content quality and pedagogical appropriateness.
The student evaluation is conducted with upper-year undergraduate students enrolled in the actual course, providing authentic feedback from learners. 

\section{Related Work}




Recent surveys have shown that LLMs are used
to assist students, teachers, and for adaptive learning~\cite{wang2024large}.
The potentials of multimodal models---models that combine text, image, and audio---are also being explored 
for STEM domains 
(e.g., in \citet{xing2024survey}).



In computing education, early work showed that code-aware models like Codex could solve most introductory programming problems~\cite{finnie2022robots}, prompting questions about assessment validity and curricular goals. More recently, \citet{denny2023conversing} explored using natural language prompts to generate CS1 solutions, while \citet{savelka2023thrilled} documented GPT-4's improved performance on higher education programming assessments, underscoring rapid capability growth. 
Recent work found that instructors are adapting their teaching based on the widespread availability of generative AI tools, such as changing assessment practices and learning objectives~\cite{prather2025beyond}.


Beyond student-facing applications, emerging work explores instructor support for content creation. \citet{sarsa2022automatic} demonstrated that LLMs could generate programming exercises and explanations, though with limitations in diversity and pedagogical appropriateness. More recently, \citet{logacheva2024evaluating} found that newer LLMs perform better at generating thematically contextualized exercises, even though the contextualization was often shallow. \citet{jordan2024need} found that LLMs can create suitable exercises in multiple languages, although exercises generated in low-resource languages such as Tamil had lower quality.
\citet{villegas2024generation} found that LLMs can be used to create culturally relevant textbooks for CS1. 
Taken together, these works show the potential and interest in leveraging LLMs in computing education. 


Slide generation, in particular, is 
under-explored. 
Early work on automatic slide generation focused on extracting content from scientific papers~\cite{sefid2019automatic}, while more recent approaches leveraged LLMs. 
\citet{xie2026slidebot} present SlideBot, a multi-agent framework that generates educational slides using specialized agents for retrieval, summarization, figure generation, and LaTeX formatting.
Other recent systems include PASS~\cite{aggarwal2025pass} for automated slide and speech generation, AutoPresent~\cite{ge2025autopresent} for designing structured visuals from scratch, and SlideCraft~\cite{rao2025slidecraft} as a context-aware slide generation agent.
These works primarily focused on system development and technical evaluation, 
rather than instructor and student experience.



Research on student perceptions of LLM-generated educational resources is limited. 
Prior work suggests that disclosure of AI assistance affects perception, e.g. in writing~\cite{li2024does}.
However, \citet{leinonen2023comparing} found LLM-generated code explanations rated comparably to student-written ones.
Similarly, \citet{henderson2025comparing} compared student perceptions of generative AI versus teacher feedback, finding differences in perceived usefulness and trustworthiness.
For slides specifically, \citet{georgiev2024exploring} examined student perceptions of instructor-created AI-assisted PowerPoint slides, finding that while students had favorable opinions of instructor AI usage, they expressed substantial concerns about slide design including structure, consistency between text and images, and typography.



\section{Textbook to Slides in a Web Software Course: Are They (GenAI Tools) Good?}
\label{sec:web-instructor}

This section explores whether and how GenAI tools could help transform chapters from an online web programming textbook into lecture slides.
This task represents a common need: instructors often maintain reading materials and generate corresponding slides. 

We used course materials from a Web Software Development (WSD) course at Aalto University in Finland: a 5 ECTS\footnote{European Credit Transfer System, 1 ECTS corresponds roughly to 27 hours of work.} course providing an introduction to contemporary web software development. The course provides a ``Full-Stack'' view, teaching students to develop client-side and server-side applications backed by a database. Early parts of the course focus on client-side development with Svelte and SvelteKit, followed by designing server-side APIs that interact with a database, which are then later integrated to the client-side functionality. 
The course instructor maintains an online interactive textbook with intertwined examples and exercises. 

We assessed various GenAI tools by producing 
slides for four topics covered in the first two weeks:
basic HTML and Svelte, sharing states, CRUD (Create, Read, Update, and Delete), and the Fetch API. 
The instructor for the course chose these topics to represent a variety of content types---some heavy on code examples, others on diagrams or conceptual explanations.
For each topic, our goal was to produce slides for a 10-15 minute lecture segment.

We explored five AI tools representing different approaches to content generation:
NotebookLM, 
M365 Copilot, 
Claude (Sonnet 4.5), 
Cursor, 
and Claude Code, 
These tools were chosen for their accessibility to computing educators and represent different points on the spectrum from automated to instructor-in-the-loop workflows.


\subsection{Methods}

Two researchers, both computing educators, collaboratively explored each tool in pair-coding sessions, iteratively developing and refining prompts. 
To mimic the style of existing lecture materials, we collected instructor-generated lecture slides from earlier course iterations.
We aimed to follow realistic instructor workflows: crafting straightforward prompts without requiring prompt engineering expertise, then refining based on what the tools produced.
After initial collaborative exploration, one researcher continued building the remaining materials for all lecture topics.


For most tools, we used the following prompt, developed iteratively through initial trials:
\begin{center}
\fbox{\parbox{0.95\linewidth}{%
        ``Please generate a set of slides from <file>. The slides are intended to be used in a 15 minute lecture segment in a 2-hour web programming lecture in a university in a web-software development course. Please include some peer-instruction exercises at the conclusion of the segment.''
    }}
\end{center}

We had initially included styling rules, but found that this degraded content quality, so we separated styling as a post-processing step for Section~\ref{sec:web-student}.
We used default settings for all tools.
For all tools, we used a separate session for each of the four chapters tested.

We provide narrative assessment of the quality of the generated materials by both the instructor and two non-instructor researchers.
We consider three factors, similar to \citet{xie2026slidebot}. 
\textbf{Factual accuracy} assessed whether the content was correct by referencing source materials, fact-checking examples, verifying conceptual explanations, and identifying hallucinations or misleading information.
\textbf{Completeness} assessed whether important concepts were omitted, whether examples were sufficient, and whether materials could stand alone or required substantial supplementation. 
\textbf{Pedagogical soundness} assessed whether generated materials aligned with established principles of computing education, including cognitive load management \cite{sweller1988cognitive,mayer2005cognitive} (presenting information in digestible units, avoiding overwhelming detail), dual coding and visual support \cite{clark1991dual,mayer2005cognitive} (using diagrams and illustrations to reinforce rather than duplicate text), and scaffolding and conceptual progression \cite{winslow1996programming,sorva2013notional} (building from concrete examples toward abstraction without abrupt jumps).

The evaluation was conducted qualitatively and collaboratively, prioritizing narrative insights for practitioners over inter-rater reliability. 
Initially, two researchers collaboratively evaluated several
generated resources, including at least two sets of slides per GenAI tool.
The researchers identified specific issues related to accuracy, completeness, and pedagogical soundness.
Through this process, we found that clarity, emphasis, and use of examples were important elements of pedagogical soundness.
Then, another researcher, who is the instructor of the web programming course, reviewed the notes, 
evaluated the remaining slides, 
and summarized the findings.

\subsection{Results}
\label{sec:instr-results}

Table~\ref{tab:web-evaluation} summarizes our assessment of the generated resources. The remaining subsections provide narrative details.
Overall, the programming-based tools, Cursor and Claude Code, produced the most compelling slides and are used in Section~\ref{sec:web-student}.

\begin{table}[t]
\centering
\begin{tabular}{p{1.75cm} p{1.5cm} p{1.8cm} p{1.7cm}}
\toprule
\textbf{GenAI Tool} & \textbf{Accuracy} & \textbf{Completeness} & \textbf{Pedagogical Soundness} \\
\midrule
NotebookLM & 
Generally
Accurate &
Very Incomplete 
& 
Very Problematic  \\ 
\midrule
M365 Copilot & 
Generally Accurate &
Incomplete & 
Problematic 
\\
\midrule
Claude & Mostly Accurate & Somewhat incomplete & Mostly acceptable \\ 
\midrule
Cursor &  Accurate & Complete 
& Strong 
\\
\midrule
Claude Code &  Accurate & Complete 
& 
Strong \\ 

\bottomrule
\end{tabular}
\caption{Evaluation of GenAI tools for generating slides from textbook materials in a Web Programming course.}
\vspace{-2em}
\label{tab:web-evaluation}
\end{table}


\subsubsection{NotebookLM}

NotebookLM is a free education-focused tool that can create multimodal learning resources~\cite{wang2024notebooklm}, particularly podcasts~\cite{ibtimes2024fake}.
At the time that this research was conducted, 
NotebookLM provided functionalities to generate ``Audio Overview'', ``Video Overview'', quizzes,
mind maps, and custom reports based on uploaded source materials;
it did not have an out-of-the-box method for generating slides.
However, as NotebookLM is advertised as a general-purpose educational tool, our students likely encounter it, so we were compelled to evaluate it.
We evaluated the slides and visual materials used in the ``Video Overview'' using the ``deep dive'' setting, generally taking $\sim$20 minutes to produce.


The visuals were \textbf{generally accurate},
though still with occasional errors like considering
hyperlink tags like \verb|<a>| to be ``interactive''.
Content was \textbf{very incomplete} and superficial, even though it was visually appealing. 
For example, the heading tag \verb|<h1>| is discussed, but other heading levels were not.
The slides were also \textbf{very problematic} pedagogically: the most obvious feature of the ``Video Overview'' itself was the long-winded introduction and motivation for the topic, also described elsewhere~\cite{gu2025when}. 
The language was often unclear and non-specific. While there were many analogies, not all of them were of high quality (e.g., of HTML as a blueprint, using a house blueprint in the title slides).
The visuals did not contain enough code examples, and when examples were present, they generally did not support comprehension.
Figures were sometimes present, but generally showed high-level processes using non-technical language.

We experimented with generating textual descriptions of the slides via the chat interface and with the
custom report generation tool.
The descriptions were comparable to other tools; however this option does not save much time for instructors.




\subsubsection{M365 Copilot}

M365 Copilot is a general-purpose GenAI assistant integrated into Microsoft 365 applications, and is available through an institutional license at one of the author's institutions. 
For M365 Copilot, we began with the prompt above, attaching
a file containing the textbook markdown source, and appending to the prompt text 
``Please generate either a PDF file or a PowerPoint file''.
The tool was able to produce a M365 PowerPoint presentation in SharePoint directly,
which we then evaluated.

The generated content was \textbf{generally accurate}.
However, some content was somewhat confusing as it missed key information: 
for example referencing textbook content (e.g., specific code example) that was not included.
Thus, the slides were \textbf{incomplete}: there were relatively few concrete examples, despite the textbook following a pattern of first introducing a concept, then code examples, and then exercises.
Much of the content seemed very surface level.
The slide design exhibited \textbf{problematic pedagogical soundness} issues: many slides had irrelevant large images that were not related to the content or helpful for understanding the concepts presented (e.g., several large images of birds). Slides also included too much text. In addition to the poor balance between image size and text, there were some minor rendering issues (e.g., related to HTML tags, and inconsistent font size).


\subsubsection{Claude}
Claude is a general-purpose LLM created by Anthropic. For this paper, we used the Pro version of Claude at a subscription rate of USD\$24 per month.
We used the same prompt above,
attaching a file containing the textbook markdown source, and 
appending to the prompt text ``Please generate either a PDF file or a PowerPoint file''.
The Chain-of-Thought output 
showed that
internally, Claude built the slides using html2pptx.
Claude iterated to ensure that the produced output fit inside the screen, and therefore took a significant amount of time (slower than Copilot).
The tool produced a downloadable PowerPoint file.

Claude's output was \textbf{mostly accurate} overall, although there were some issues caused by HTML rendering. For example, a slide on key HTML elements had bulleted text explanations where each HTML tag was missing (e.g. `` -- Declares document type''), likely due to issues rendering the characters \verb|<| and \verb|>|. Similarly, another slide showed that both unordered and ordered lists would result in bulleted item lists, which is incorrect. 
In terms of \textbf{completeness},
Claude also did not provide as many examples (e.g., of the DOM's tree-like data structure). 
The instructor would need to fill in considerable content to make the slides classroom ready. 
The \textbf{pedagogical soundness} was somewhat mixed but mostly acceptable; although the slides followed the materials in general, some slides jumped directly from high-level descriptions to benefits, missing concrete examples that one would typically show in a classroom. On the positive side, Claude also identified content to highlight in slides, and the textual content was mostly well balanced.


\subsubsection{Cursor}
Cursor is a coding assistant and AI-powered code editor built on VS Code. For this paper, we used the Pro version of Cursor (1.7) at a subscription rate of USD\$20/month. 
Since Cursor is primarily an integrated development environment (IDE) intended to produce code,
we asked Cursor to produce source materials that could be built into a set of slides.
We appended to the prompt in Section~\ref{sec:web-instructor} the following instruction:
``Use Quarto to generate the slides. Include all makefiles, etc., so that the slides can be built.  Please put the results in the <folder>''.
Quarto was chosen since it is frequently used by computing educators, because one of the authors was familiar with it, and because another author's name rhymed with it.
In addition to the slides, 
Cursor generated presenter notes for instructors without prompting.

These slides were generally \textbf{accurate} and \textbf{complete}. During the evaluation, one of the questions that the evaluators raised was whether the slides were \textit{too} complete; however, instructors could easily discard content. 
There were some minor inaccuracies that could prompt misconceptions, e.g., where the JavaScript and Svelte code was shown side-by-side but were not equivalent.
The content was, \textbf{pedagogically, fairly strong}. Many examples were used, e.g., providing multiple examples for CRUD. 
The slides leveraged parallelism, e.g., ordered vs unordered lists were rendered side by side. Diagrams were used, most of which were from the source materials.
There were some terms (like ``RESTful API'') that were used before they were introduced, though generally new terms were bolded and emphasized. 
The slides also had good progression and the content was mostly presented in a step-by-step fashion, where items would appear on slides one by one, which was the style that the instructor preferred and used. At the same time, some extra styling would be required: some slides did not fit inside the frame, and the font size in code examples was often too small.


\subsubsection{Claude Code}

Claude Code is a command line tool for agentic coding, available with the Claude Pro subscription.
The interaction affordances were similar to Cursor, and  Claude Code (version 2.0.29) also decomposed the prompt into multiple sub-tasks. The same prompts and methods were used as Cursor.

Claude Code's output was mostly\textbf{ accurate}, with only minor issues such as a slide captioned ``text paragraphs'' also featuring headings without extra information that would guide the reader. In terms of completeness, Claude Code's slides were \textbf{mostly complete}, with some examples of missing relevant information. 
In some instances, similar to Cursor, the output was ``too complete'', meaning that the instructor would need to decide what content to keep. 
Regarding \textbf{pedagogical soundness}, similar to Cursor, Claude Code's slides had good logical progression, with theory intertwined with code examples. However, some slides contained content in `iframe'-elements that required scrolling,
and code examples were missing syntax highlighting.





\section{Student Perception of GenAI Slides: \\ Can Students Tell?} 
\label{sec:web-student}


Based on the results in Section~\ref{sec:web-instructor},
we were confident that with minor modifications, the materials
generated by Cursor and Claude Code can be used
without negatively impacting student experience.
All researchers agreed that Cursor and Claude Code generated slides were superior
to the remaining tools, but disagreed on which was best; the web development course instructor
preferred Cursor.

We thus evaluated the student perceptions of these materials
in an instance of the Web Software Development course.
This course included five 2-hour lectures.
The lectures were organized using a peer instruction style; they were divided into segments with slides that focus on specific core areas. 
During each segment, the instructor complemented the slides with
live-coding, and provided additional insights and examples.
The instructor has taught web software development-related courses for over a decade, and has prior web development experience from the software industry.
At the end of each segment, there were multiple-choice questions and possibly class discussions, depending on student performance in the multiple-choice questions.
Active attendance (e.g., actively responding to the multiple-choice questions) in lectures provided a small amount of extra points (up to 5\% of overall grade). 
A total of 88 students enrolled in the course in the fall of 2025, and 51 attended the first lecture.  

\subsection{Methods}

We evaluated a combination of GenAI and human-generated slides.
For weeks 1 and 2, we used the 4 slide segments generated by Cursor from Section~\ref{sec:web-instructor}.
For weeks 3-5, we generated 6 additional slide segments using Claude Code.
Slides for the remaining segments were created by the instructor without GenAI.

\begin{table*}[ht]
\centering
\caption{Overall quality and human vs.\ AI guesses by segment on a 7-point Likert scale. For quality, 1 = very bad, 7 = excellent,  for human vs.\ AI, 1 = definitely human, 7 = definitely AI, and for source AI-C = Cursor, AI-CC = Claude Code. Only students who reported attending the lecture are included. Also Spearman correlation between overall quality and human vs.\ AI guess.}
\label{tab:combined_stats}
\small
\begin{tabular}{p{0.41cm}lll|cccc|cccc|cc}
\toprule
\multicolumn{4}{c}{\textbf{Lecture Segment}} &
\multicolumn{4}{c}{\textbf{Overall quality}} &
\multicolumn{4}{c}{\textbf{Human vs.\ AI guesses}} &
\multicolumn{2}{c}{\textbf{Correlation}} \\
\cmidrule(lr){1-4} \cmidrule(lr){5-8} \cmidrule(lr){9-12} \cmidrule(lr){13-14}
 & \textbf{Topic} & \textbf{Source} & \textbf{n}
 & \textbf{Mean} & \textbf{Median} & \textbf{SD} & \textbf{Range}
 & \textbf{Mean} & \textbf{Median} & \textbf{SD} & \textbf{Range}
 & \textbf{$\rho$} & \textbf{$p$} \\
\midrule
1-1 & Practicalities & Human & 22 & 5.5 & 6.0 & 1.2 & [3, 7] & 3.5 & 3.0 & 1.7 & [1, 7] & 0.42 & 0.05 \\
1-2 & HTML & AI-C & 18 & 5.5 & 5.5 & 1.1 & [3, 7] & 3.6 & 3.5 & 1.5 & [1, 7] & -0.48 & 0.03 \\
1-3 & JS and Components & Human & 16 & 5.1 & 5.0 & 0.8 & [4, 7] & 3.5 & 4.0 & 1.6 & [1, 6] & -0.39 & 0.12 \\
1-4 & Pages and Components & Human & 16 & 5.0 & 5.0 & 1.3 & [2, 7] & 3.8 & 4.0 & 1.3 & [1, 6] & -0.46 & 0.06 \\
1-5 & Reactivity & Human & 20 & 4.4 & 4.0 & 1.5 & [2, 7] & 4.0 & 4.0 & 2.0 & [1, 7] & -0.41 & 0.07 \\
1-6 & Shared State & AI-C & 12 & 4.7 & 5.0 & 0.8 & [4, 6] & 4.7 & 4.0 & 1.5 & [2, 7] & 0.49 & 0.09 \\
\midrule
2-1 & HTTP & Human & 28 & 5.3 & 5.5 & 0.9 & [3, 7] & 3.0 & 3.0 & 1.5 & [1, 6] & -0.39 & 0.04 \\
2-2 & Hono & Human & 28 & 5.3 & 5.5 & 1.3 & [3, 7] & 3.6 & 4.0 & 1.4 & [1, 6] & -0.10 & 0.62 \\
2-3 & Building an API & Human & 28 & 5.1 & 5.0 & 0.9 & [3, 7] & 3.7 & 4.0 & 1.2 & [2, 6] & -0.07 & 0.72 \\
2-4 & CRUD & AI-C & 27 & 4.8 & 5.0 & 1.0 & [2, 7] & 4.4 & 4.0 & 1.6 & [2, 7] & -0.50 & $<0.01$ \\
2-5 & Fetch API & AI-C & 26 & 4.8 & 5.0 & 0.9 & [3, 7] & 4.0 & 4.0 & 1.5 & [1, 7] & -0.31 & 0.12 \\
\midrule
3-1 & Auth x2, Storing Passwords & Human & 21 & 5.0 & 5.0 & 0.9 & [4, 7] & 3.3 & 3.0 & 1.7 & [1, 7] & -0.35 & 0.12 \\
3-2 & Tracking Users & Human & 22 & 5.0 & 5.0 & 1.1 & [3, 7] & 3.4 & 4.0 & 1.3 & [1, 6] & -0.51 & 0.02 \\
3-3 & Client-side User Management & Human & 22 & 4.9 & 5.0 & 1.2 & [3, 7] & 3.4 & 4.0 & 1.4 & [1, 5] & -0.40 & 0.07 \\
3-5 & Web Security Essentials & AI-CC & 22 & 5.3 & 5.0 & 0.9 & [4, 7] & 3.1 & 3.0 & 1.4 & [1, 6] & -0.63 & $<0.01$ \\
\midrule
4-1 & Cascading Style Sheets & AI-CC & 23 & 5.0 & 5.0 & 0.9 & [3, 7] & 3.5 & 4.0 & 1.3 & [1, 6] & -0.13 & 0.57 \\
4-2 & Styling with Tailwind CSS and Skeleton & Human & 21 & 4.9 & 5.0 & 1.0 & [4, 7] & 3.8 & 4.0 & 1.4 & [1, 7] & -0.37 & 0.10 \\
4-3 & Responsive Web Design & AI-CC & 22 & 5.0 & 5.0 & 1.0 & [4, 7] & 3.8 & 4.0 & 1.4 & [1, 7] & -0.37 & 0.10 \\
4-4 & Web Accessibility and WCAG & AI-CC & 22 & 5.0 & 5.0 & 0.9 & [3, 7] & 3.6 & 4.0 & 1.6 & [1, 6] & -0.43 & 0.06 \\
\midrule
5-1 & Unit Testing etc & AI-CC & 21 & 5.0 & 5.0 & 1.0 & [3, 7] & 3.5 & 3.5 & 1.5 & [1, 6] & -0.24 & 0.30 \\
5-2 & End-to-End Testing & AI-CC & 21 & 6.0 & 6.0 & 1.1 & [4, 7] & 3.5 & 3.0 & 2.2 & [1, 7] & -0.26 & 0.32 \\
5-3 & Evolution of Web Development & Human & 16 & 5.2 & 6.0 & 1.1 & [4, 7] & 3.5 & 3.0 & 2.2 & [1, 7] & -0.26 & 0.32 \\
\midrule
\multicolumn{2}{l}{Overall (AI-C / Cursor)} & AI-C & 83
 & 5.0 & 5.0 & 1.0 & [2, 7]
 & 4.1 & 4.0 & 1.6 & [1, 7]
 & -0.32 & $<0.01$ \\
\multicolumn{2}{l}{Overall (AI-CC / Claude Code)} & AI-CC & 116
 & 5.1 & 5.0 & 0.9 & [3, 7]
 & 3.5 & 4.0 & 1.5 & [1, 7]
 & -0.32 & $<0.01$ \\
\midrule
\multicolumn{2}{l}{Overall (AI)} & AI & 199 & 5.1 & 5.0 & 0.9 & [2, 7] & 3.8 & 4.0 & 1.5 & [1, 7] & -0.33 & $<0.01$ \\
\multicolumn{2}{l}{Overall (human)} & Human & 260 & 5.1 & 5.0 & 1.1 & [2, 7] & 3.5 & 4.0 & 1.5 & [1, 7] & -0.24 & $<0.01$ \\
\midrule
\multicolumn{2}{l}{Overall (all)} & - & 459 & 5.1 & 5.0 & 1.0 & [2, 7] & 3.6 & 4.0 & 1.5 & [1, 7] & -0.28 & $<0.01$ \\
\bottomrule
\end{tabular}
\end{table*}

To ensure blinding, we matched the GenAI slides to the instructor's visual style, fixing formatting issues through iteration.
Some of this work was done directly in Cursor, using the following prompt:

    \begin{center}
    \fbox{\parbox{0.95\linewidth}{%
        ``I would like to update <path\_to\_slides> to `look and feel' like the instructor's slides. Here are some example images showing the rendered example slides.
        The instructor likes to use Arial font for the main text.
        The instructor also likes to use red text with `Coming Soon' font, slightly slanted, to explain things in the code or bullets. There are transitions associated with these red text segments (they do not appear all at once).
        Any code fragments uses Consolas font. The styling used is consistent with those used in VSCode by default, but with the white background.
        I would like you to do this update by creating a new file called <new\_path>. Please keep the content as similar as possible to the current version, though I understand that some content will need to change for the `red text' to make sense.''
    }}

    \end{center}
We then followed up with ``Could you render slides\_styled into a PPTX file?''
Styling was still challenging for the tool. We note that the GenAI styled presentations were inconsistent and contained issues.
For example, the instructor used red overlay text in a font resembling handwriting to annotate key ideas and provide commentary, essentially drawing on the signaling principle to help learners understand the content~\cite{mayer2017using}. 
The signaling, including styling, font, and transitions, were difficult for GenAI tools to emulate. 
The instructor fixed the style before using the materials. 

Additionally, when adding the AI-generated slides to the slide deck of the lectures, the instructor was allowed to delete content, e.g., entire slides, examples, or other content (e.g., the ``learning objectives'' slides). 
We believe deletion is a reasonable, efficient, and realistic change that instructors would make when using GenAI tools. 
These adjustments required little effort, and making them helped the instructor familiarize themselves with the slides.

Students were informed that lectures might contain GenAI-generated content but not told which segments or how many.
After each segment in the lectures, students completed a brief survey rating the segment quality on a 7-point Likert scale\footnote{\textit{Overall, on a scale from 1 = Very bad to 7 = Excellent, how would you rate the slides in the previous segment?}} and guessing whether it was GenAI-generated\footnote{\textit{On a scale from 1 = Definitely Human to 7 = Definitely AI, what’s your best guess: were the slides in the previous segment AI-made or human-made?}}. Students were instructed to focus their evaluation only on the slides, not other content of the segment. Participating in the survey was voluntary and did not contribute towards the points that students received from being active in the lectures\footnote{In the first lecture, the software used for collecting the ratings failed to store a part of the student responses, while all student responses from the second lecture onwards were stored.}.
The study had ethics approval from the university's ethics committee.

We note two limitations to this approach:
First, individual students may have contributed different numbers of ratings\footnote{This violates the independence assumption of the Mann-Whitney U test, and some students could influence results more than others.}. 
Second, despite students being told to focus on the slides in their evaluation, the peer instruction ratings might reflect not only the quality of the slides but also other aspects of the instructional segment, such as live coding or verbal explanations. 

We calculated the mean, median, standard deviation and range of students' responses to the two Likert-scale questions per segment and overall. We examined if there were any differences between AI- and human-made slides using a Mann-Whitney U test (since the data was ordinal). In addition, we analyzed the correlation between the two questions to see if students' perceptions of the slides correlates with their guess of whether the slides were AI-generated.

\subsection{Results}



The means, medians, ranges, and correlations of the analysis can be seen in Table~\ref{tab:combined_stats}. We found that the quality of the AI- and human-made slides was similar: the median for both was 5.0. This is supported by the Mann-Whitney U test results (U = 25600, p = 0.84), which indicate we failed to find evidence that the slides were of different quality. While the median for the guesses for both AI and human slides was 4.0, students seemed not to be able to guess the source (AI vs. human) of the slides based on the Mann-Whitney U test results (U = 28171.5, p = 0.1). Essentially, this suggests that \textbf{the quality of the AI slides is similar to that of human slides, and students cannot reliably distinguish the source of the slides.}

We also see a weak, negative correlation between perceived quality and guess on whether the slides were produced by AI or a human (-0.28, p <0.01). This suggests that, even though the actual quality of the slides between the AI and human was similar, \textbf{students associate poor quality with AI as the source}. 
This mirrors findings that humans rate writing quality as being lower when its source is revealed to be AI~\cite{li2024does}.

Looking at the biggest mismatches in guesses (i.e., human slides most rated as AI and vice versa), we found some potential explanations for students' ratings. Slide set 1-5, which was human-created, had the highest average score overall for AI being the source among the human-created slides. This was likely due to a rendering issue where the code on the slide did not display correctly. Similarly, slide set 3-5 had the lowest average score for being AI among AI slides, and second lowest overall even though it was created by Claude Code. A likely explanation is that the slide set included the XKCD comic ``Exploits of a Mom''\footnote{\url{https://xkcd.com/327/}} about ``little Bobby Tables'', which probably caused students to think the slides were by a human.

\section{Discussion and Recommendations}

We were positively surprised by the quality of the materials produced by 
Cursor and Claude Code. 
While edits were required, 
the instructor of the web programming course found  lecturing with the slides similar to using one's own slides or a colleague's slides.

We purposefully included different types of GenAI tools but found that coding assistants worked best, even for tasks that are not traditional ``programming'' tasks. 
We suggest that \textbf{computing instructors think of themselves as programmers first}. Similar to programmers working in the industry, instructors can work iteratively with tools, prompting and re-prompting, highlighting issues, etc.~\cite{khojah2024beyond}. 
This includes building slides in a tool like Quarto to generate HTML, rather than PowerPoint, so that coding assistants can more easily work with textual representations of the materials.
Thinking of oneself as a programmer also means leveraging existing documentation and tooling meant for software engineers: for example, we did not write an \verb|AGENTS.md|, provide custom configuration to the coding assistants, 
or use parallel agents.
We suspect that more advanced use could provide additional time-saving.
Coding assistants require programming knowledge, and are more accessible to computing instructors than those in other fields.

Rather than expecting a perfect first-draft,
\textbf{willingness to iterate was key in using GenAI tools.}
The drafts were imperfect, and the GenAI tools were particularly poor at styling the content appropriately. 
The drafts also still contained inaccuracies despite recent improvements in the underlying models~\cite{OpenAI2025GPT5SystemCard}. 
Moreover, it is possible for the generated material to duplicate copyrighted material. However, in our spot checking, we did not find content that was verbatim copied from another source. 
All the content we explicitly provided to the models was originally created by the instructor who participated in this study. 
We encourage instructors to check their institution's guidelines on GenAI use.

We also believe that \textbf{GenAI can be effectively leveraged without 
hurting student experience when the instructor remains in the loop}. 
In Section~\ref{sec:web-student},
we found no statistically significant difference between students' rating of GenAI vs. instructor slides, and their ratings of the qualities were the same (mean rating of 5.1). 
The statistically significant negative correlation between quality and guess suggests that the AI would have been rated even higher in quality if it was more similar to instructor slides in style.
This, again, emphasizes iteration: with more editing of just the style (which we believe is not time-consuming), perceived quality might be higher.

We emphasize that our results are a snapshot in time in late 2025,
and models and tools improve constantly.
For example, during the writing of the paper, Cursor 2.0 was released.
Shortly after, NotebookLM added support for slide deck generation.
Our experience is still useful to share, 
to document GenAI capabilities at a snapshot in time,
while presenting timely insights for practitioners.

We also emphasize that the study was run at a single institution and course, so the results might not generalize beyond this context. For example, it could be possible that coding tools worked better because the context was programming related. We also used Claude Code and Cursor on different weeks of the course, which could have affected the results, and we did not measure effects on learning.

\section{Conclusion}

We described a case study of using generative AI tools to support the creation of lecture slides based on existing course e-book materials. Our goal was to report our holistic initial experiences, and not to conduct a research study. We envision future work that evaluates these ideas more rigorously. We found that using generative AI tools could save significant instructor time, with coding assistant tools being particularly powerful. 
We also found no statistically significant differences in student ratings of the quality between AI- and human-generated slides, and students seemed not to be able to guess which slides were created by AI. These findings highlight promising opportunities for integrating generative AI into instructional design workflows and call for further research on how educators can best harness such tools responsibly and effectively.

\begin{acks}
This work was supported by Research Council of Finland grants \#356114 and \#367787. Generative AI was used to assist with the writing and polishing of the text in this paper. All GenAI-generated or polished text was reviewed by the (human) authors and the human authors take full responsibility for the text.
\end{acks}

\balance
\bibliographystyle{ACM-Reference-Format}
\bibliography{references}


\end{document}